\documentclass[10pt,conference]{IEEEtran}
\IEEEoverridecommandlockouts
\usepackage{microtype}
\usepackage{graphicx}
\usepackage{subfigure}
\usepackage{booktabs} 
\usepackage{color}
\usepackage{amsmath}
\usepackage{algorithmic}
\usepackage[noadjust]{cite}
\usepackage{float}
\usepackage{mathtools}
\usepackage{comment}
\usepackage{xfrac}
\usepackage{bbm}
\usepackage{multirow}
\usepackage[cmintegrals]{newtxmath}
\usepackage{caption}
\usepackage{titlesec}

\begin{document}

\title{Fair Kernel Regression via 
Fair Feature Embedding in Kernel Space}

\author{\IEEEauthorblockN{Austin Okray}
        \IEEEauthorblockA{Department of Computer Science\\
        University of Wyoming\\
        Laramie, Wyoming\\
        Email: aokray@uwyo.edu}
    \and
        \IEEEauthorblockN{Hui Hu}
    \IEEEauthorblockA{Department of Computer Science\\
        University of Wyoming\\
        Laramie, Wyoming\\
        Email: hhu1@uwyo.edu}
    \and
    \IEEEauthorblockN{Chao Lan}
    \IEEEauthorblockA{Department of Computer Science\\
        University of Wyoming\\
        Laramie, Wyoming\\
        Email: clan@uwyo.edu}
}

\maketitle


\begin{abstract}
In recent years, there have been significant 
efforts on mitigating unethical demographic biases 
in machine learning methods. 
However, very little work is done for kernel methods. 
In this paper, we propose a novel fair kernel 
regression method via fair feature 
embedding (FKR-F$^2$E) in kernel space. 
Motivated by prior works feature processing 
for fair learning and feature selection for 
kernel methods, we propose to learn 
fair feature embeddings in kernel space, 
where the demographic discrepancy of 
feature distributions is minimized. 
Through experiments on three public real-world 
data sets, we show the proposed FKR-F$^2$E achieves 
significantly lower prediction disparity compared with 
the state-of-the-art fair kernel regression method 
and several other baseline methods. 
\end{abstract}


\section{Introduction}
\label{introduction}

In recent years, we've witnessed 
a tremendous 
growth of machine learning applications in 
real-world problems that 
have immediate impacts 
on peoples' lives. However, standardly learned models 
can have unethical predictive biases against minority 
peoples; e.g., in recidivism prediction, a
commercialized model has significant bias against innocent black defendants \cite{angwin2016compas}; other biases 
are found in hiring \cite{hoffman2017discretion}, 
facial verification \cite{klare2012face}, violence 
risk assessment in prison \cite{cunningham2006RASP}, 
etc.

How to learn fair models has become a 
significant research topic \cite{press2016preparing},  
and many methods have been proposed
\cite{zemel2013LFR, feldman2015certifying,
Calders2013ControllingAE, kamishima2012fair, dwork2012fairness, McNamara2017ProvablyFR, samadi2018fpca}. They typically
sacrifice certain prediction accuracy for 
improving prediction fairness, bound to 
the accuracy-fairness tradeoff. 

A promising direction is fair kernel learning
\cite{perez2017fair, Olfat2019ConvexFF}. 
By constructing sufficiently complex hypothesis 
spaces, they are more likely to learn a model 
that can achieve an efficient accuracy-fairness trade-off. 
However, this direction is sparsely explored so far. 
A notable work is fair kernel regression 
\cite{perez2017fair}, which penalizes 
a model's predictive bias in kernel space. 

In this paper, we propose a novel fair kernel regression 
method that learns fair feature embeddings (FKR-F$^2$E) in 
the kernel space. It is motivated by the work of 
Feldman et al \cite{feldman2015certifying}, 
which shows that in a properly transformed data space 
where different demographic groups have similar 
feature distributions, a standardly learned prediction 
model will be naturally fair. 
We thus seek for such a fair transformation in 
the kernel space. 

A major challenge is that kernel space is often 
implicit, making it hard to find explicit fair transformations 
therein. To tackle the problem, we borrow ideas 
from Cao et al \cite{cao2007basevec}, which learns 
feature embeddings in the kernel space for 
feature selection. 
Specifically, we propose to learn fair 
feature embeddings in the kernel space, 
such that different demographic groups 
have similar embedded feature distributions. 
We propose to measure similarity using 
mean discrepancy \cite{gretton2012mmd}.

Through experiments on three real-world data sets, 
we show the proposed FKR-F$^2$E achieves significantly 
lower prediction bias than the existing fair kernel 
regression method as well as several non-kernel fair 
learning methods, without sacrificing a significant 
amount of prediction accuracy. 

The rest of the paper is organized as follows: 
in Section II, we revisit related works; 
in Section III, we present the proposed method; 
in Section IV, experimental results are presented 
and discussed; our conclusion is in Section V. 

\subsection{Notations and Assumptions}

To facilitate discussions in related 
work, we introduce some notations here. 
We describe an instance using a triple 
$(x,s,y)$, where $x$ is a feature vector, 
$s$ is a protected demographic  
(e.g. gender, race) and $y$ is label. 
Assume $s$ is contained in $x$. 

Similar to prior studies, 
we assume $s$ is binary. 
Let there be $n$ instances in the 
training set, among which $n_{u}$ belong 
to the unprotected group (s = 0) and 
$n_{p}$ belong to the protected group (s = 1). 
Without loss of generality, we assume 
the instances are ordered such that 
the first $n_{u}$ ones 
$x_{1}, \ldots, x_{n_{u}}$ are unprotected 
and the rest $x_{n_{u}+1}, \ldots, x_{n}$ 
are protected. 

For kernel methods, let $\phi(\cdot)$ be the 
feature mapping function, and $f$ be a prediction 
model mapping from $\phi(x)$ to $y$.

\section{Related Work}

\subsection{Fair Kernel Regression}

Perez-Suay et al \cite{perez2017fair} propose 
a fair kernel regression method, which directly 
extends the linear fair learning method \cite{kamishima2012fair} to kernel space.
Specifically, it minimizes prediction loss 
while additionally penalizing the correlation 
between model prediction and demographic feature 
in the kernel space as:
\begin{align}
\begin{split}
& \min_{f} \sum_{i=1}^{n} [ f(\phi(x_{i})) - y_{i} ]^{2} 
+ \mu \sum_{i=1}^{n} (\bar{f}(x_{i}) \cdot \bar{s}_{i})
+ \lambda \Omega(f),
\end{split}    
\end{align}
where the second term measures predictive bias 
as the correlation between model prediction 
and the demographic feature, and $\bar{f}(x)$ 
and $\bar{s}$ are centered variables;  
the last term measures model complexity; 
$\mu$ and $\lambda$ are hyper-parameters. 
Based on the Representer Theorem 
that $f$ is a linear combination of $\phi(x_{i})$'s, 
task (1) admits an analytic solution for 
the linear coefficients. 

Perez et al's method adopts the regularization 
approach in fair learning, which 
penalizes predictive bias during 
learning (e.g., \cite{Calders2013ControllingAE, kamishima2012fair}).
In this paper, we adopt another popular
approach which first 
constructs a fair feature space and 
then builds a standard model in it 
(e.g., \cite{zemel2013LFR, feldman2015certifying, 
samadi2018fpca}). In experiments, we show 
our method can achieve higher prediction 
fairness.  

\subsection{Fair Feature Learning and Mean Discrepancy}

An effective approach to learn fair 
models is to first construct a fair 
feature space and then learn a 
standard model in it (e.g., \cite{zemel2013LFR,feldman2015certifying}). 
A fair feature space is one where 
feature distributions of different 
demographic groups are similar, e.g., 
different groups have similar CDF's 
of the new features \cite{feldman2015certifying}, 
or the statistical dependence 
between the new features and the 
demographic feature is low \cite{zemel2013LFR}. 

In this paper, we develop a new fair kernel 
regression based on the idea of fair feature 
learning. Unlike previous studies, 
we measure feature 
similarity using mean discrepancy 
\cite{gretton2012mmd}. MD measures 
distance between distributions and 
is widely used in machine learning \cite{huang2007correcting,pan2008transfer,gretton2009covariate}. 
Let $x_{1}, \ldots, x_{n}$ and 
$z_{1}, \ldots, z_{m}$ be two 
sets generated from 
distributions $P_{x}$ and $P_{z}$ 
respectively. MD estimates the 
distance between $P_{x}$ and $P_{z}$ as
\begin{equation}
MD(P_{x},P_{z}) 
= \left|\left| \frac{1}{n} \sum_{i=1}^{n} 
\phi(x_{i}) - \frac{1}{m} \sum_{j=1}^{m} 
\phi(z_{j}) \right|\right|^2.
\end{equation}

A technical challenge is that previous fair 
feature learning approaches assume the 
feature space is explicit and then
modify it to obtain a fairer space. 
In kernel methods, however, the feature space 
of $\phi(x)$ is implicit. To tackle this 
issue, we propose to construct an explicit 
fair feature space for $\phi(x)$, by learning 
fair feature embedding functions in the kernel space. 
This approach is motivated by the literature 
of feature selection in kernel methods 
(e.g., \cite{grandvalet2003scaling,cao2007basevec}). 

\subsection{Feature Selection in Kernel Space}

Feature selection is a common practice for improving 
the robustness and interpretability of machine learning 
models \cite{Tang2014FeatureSF}. However, its practice in kernel methods 
is not easy, since there is not an explicit feature
representation in kernel space. 
Only a few approaches are proposed, e.g. \cite{grandvalet2003scaling,yang2016modelfree,cao2007basevec}. 



Our study is motivated by Cao et al \cite{cao2007basevec}. 
They propose to learn explicit feature representation in 
the kernel space, by learning feature embedding function
$\eta$. 
They show the optimal function is  
a linear combination of training instances, 
and learn such functions by standard 
methods such as KPCA \cite{scholkopf1997kernel}. 
After that, instance $\phi(x)$ 
is mapped onto $\eta$ to obtain an explicit feature 
representation on which feature selection 
is performed. 

Motivated by Cao et al's approach, we propose 
to learn feature embedding functions that are 
fair, e.g.., different demographic groups have similar 
distributions in the embedded space. As explained 
in the previous subsection, similarity is measured by 
mean discrepancy.

\section{Fair Kernel Regression via Learning 
Fair Feature Embeddings in Kernel Space (FKR-F$^2$E)}

In this section, we present the proposed fair 
kernel regression via learning fair feature 
embeddings in kernel space (FKR-F$^2$E). 
Recall an individual is $(x,s,y)$, where 
$x$ is feature vector, $s$ is binary 
demographic feature and $y$ is label. 
There are $n$ training instances, 
where $x_{1}, \ldots, x_{n_{u}}$ 
are from the unprotected group and 
$x_{n_{u}+1}, \ldots, x_{n}$ are from 
the protected group. 

Our proposed method works in two steps: (i) 
learn fair feature embeddings in kernel space; 
(ii) build a standard regression model 
based on the embedded features. 

\subsection*{Step 1. Learn Fair Feature Embeddings 
in Kernel Space}

Our goal is to learn an explicit and fair feature 
representation for $\phi(x)$. 
To that end, we propose to learn a fair feature 
embedding function $\eta$, such that in the embedded 
space, the mean discrepancy between the protected 
group and unprotected group is minimized: 
\begin{equation}
\label{eq:obj}
\min_{\eta}  \left|\left| 
\frac{1}{n_u} \sum_{i=1}^{n_u} \langle \phi (x_i), 
\eta \rangle - \frac{1}{n_p} \sum_{i=n_{u}+1}^{n} 
\langle \phi(x_i), \eta \rangle \right|\right|^2.
\end{equation}
Problem (3) cannot be directly solved since 
there is no explicit representation of $\phi(x)$.  
Motivated by Cao et al \cite{cao2007basevec}, we 
assume the optimal $\eta$ is a linear combination 
of training instances: 
\begin{equation}
\label{eq:const1}
\eta = {\sum}_{i=1}^{n} \alpha_{i} \phi(x_{i}). 
\end{equation}
To avoid overfitting, we further assume $\eta$ 
has a unit norm: 
\begin{equation}
\label{eq:const2}
    ||\eta||^{2} = 1. 
\end{equation}
Solving (\ref{eq:obj}) under constraints 
(\ref{eq:const1}) and (\ref{eq:const2}), 
we have that\footnote{Detailed arguments are 
in Appendix A.} 
\begin{equation}
\label{eq:solutioneta}
    \left(\frac{1}{n_u^2}K_u^T K_u - \frac{2}{n_u n_p}K_u^T \mathbf{1}_u \mathbf{1}_p^T K_p + \frac{1}{n_p^2} K_p^T 
    K_p \right) \alpha = \lambda K \alpha, 
\end{equation}
where $\alpha = [\alpha_{1}, \ldots, \alpha_{n}]^T$ is 
the vector of unknown parameters and $\lambda$ is an 
eigenvalue;  $K$ is a standard
$n$-by-$n$ Gram matrix of all instances; 
$K_{u}$ is $n_u$-by-$n$ and $K_{p}$ is 
$n_p$-by-$n$ satisfying 
\begin{equation}
K = \begin{bmatrix}
    K_u \\
    K_{p}
    \end{bmatrix}
\end{equation}

Formula (\ref{eq:solutioneta}) is a generalized 
eigenproblem, and $\alpha$ is the least generalized 
eigenvector. After $\alpha$ is solved, we obtain 
the first explicit and fair feature of $\phi(x)$ 
in the kernel space as 
\begin{equation}
\langle \phi(x), \eta \rangle = 
{\sum}_{i=1}^{n} \alpha_{i} k(x, x_{i}). 
\end{equation}

The above analysis gives the first fair 
feature embedding function $\eta$ in the kernel space. 
Now we present how to obtain the second $\eta'$, 
and the rest can be derived in similar fashions. 

The second optimal embedding $\eta'$ is obtained 
in a similar fashion as $\eta$, with an additional 
constraint that it should be orthogonal to the 
previously obtained embeddings: 
\begin{equation}
\label{eq:2ndfunction}
    \begin{aligned}
    & \min_{\eta'} \left|\left| 
\frac{1}{n_u} \sum_{i=1}^{n_u} \langle \phi (x_i), 
\eta' \rangle - \frac{1}{n_p} \sum_{i=n_{u}+1}^{n} 
\langle \phi(x_i), \eta' \rangle \right|\right|^2 \\[.5em]
    & s.t.\ \eta' = {\sum}_{i=1}^{n} \alpha'_{i} \phi(x_{i}), 
    \quad ||\eta'||^{2} = 1, \quad \eta^{T} \eta' = 0. 
    \end{aligned}
\end{equation}
Solving (\ref{eq:2ndfunction}) shows that $\alpha' = [\alpha_{1}', \ldots, \alpha_{n}']^T$
is the second least generalized 
eigenvector of the same eigenproblem (\ref{eq:solutioneta})\footnote{Detailed arguments 
are in Appendix B.} . 

By similar arguments, we can show the linear coefficients 
of $k$ optimal fair embeddings $\eta_{1}, \ldots, 
\eta_{k}$ are the least $k$  generalized eigenvectors 
of the eigenproblem (\ref{eq:solutioneta}). 

After that, we obtain a k-dimensional explicit fair 
feature representation of $\phi$ in the kernel space, 
i.e., 
\begin{equation}
\phi_{FS}(x) = 
[\langle \phi(x), \eta_{1} \rangle, 
\ldots, \langle \phi(x), \eta_{k} \rangle]^{T}. 
\end{equation}
\subsection*{Step 2. Learn a Standard Regression 
Model on $\phi_{FS}(x)$}

Given an explicit fair feature representation 
$\phi_{FS}(x)$, we learn a standard regression 
model based on it. 
Let $x_{1}, \ldots, x_{n}$ be $n$ training instances. 
One can easily verify that 
\begin{equation}
\phi_{FS}(x_{i}) = 
[\langle \phi(x_{i}), \eta_{1} \rangle, 
\ldots, \langle \phi(x_{i}), \eta_{k} \rangle]^{T}
= (K_{:i}^{T} A)^T
\end{equation}
where $K_{:i}$ is the $i^{th}$ column Gram matrix $K$, 
and $A$ is an $n$-by-$k$ matrix with column j being 
the linear coefficient vector of $\eta_{j}$
(e.g., the first column is $\alpha$ and 
the second column is $\alpha'$). 

Then, one can obtain an $n$-by-$k$ training 
sample matrix
\begin{equation}
    X_{FS} =
    \begin{bmatrix}
    (\phi_{FS}(x_{1}))^T \\ \vdots\\ (\phi_{FS}(x_{n}))^T
    \end{bmatrix} 
    = 
    \begin{bmatrix}
    K_{:1}^T A\\ \vdots \\ K_{:n}^T A
    \end{bmatrix} 
    = K^TA = KA.
\end{equation}
Now, we learn a regression model $\beta \in 
\mathbb{R}^{k}$ on $X_{FS}$ by 
\begin{equation}
\min_{\beta}\ ||X_{FS} \cdot \beta - Y||^{2} + 
\gamma ||\beta||^{2},
\end{equation}
where $\gamma$ is a regularization coefficient. 

For any testing instance $z$, we first 
compute its explicit fair feature representation 
\begin{equation}
\phi_{FS}(z) = [\langle \phi(z), \eta_{1} \rangle, 
\ldots, \langle \phi(z), \eta_{k} \rangle]^{T},     
\end{equation}
and then compute its prediction as
\begin{equation}
\hat{y} = \phi_{FS}(z)^{T} \beta. 
\end{equation}
For classification tasks, one can simply threshold $\hat{y}$.

\section{Experiment}

\subsection{Data Sets}

We experimented on three public data sets,
namely, the Credit Default data set\footnote{https://archive.ics.uci.edu/ml/datasets/default+of+credit+card+clients}, 
the Community Crime data set\footnote{http://archive.ics.uci.edu/ml/datasets/communities+and+crime}, 
and the COMPAS data set\footnote{https://github.com/propublica/compas-analysis}. 

The original Credit Default data set contains 30,000 individuals described by 23 attributes. 
We treated `education level' as the sensitive variable, and binarized 
it into higher education and lower education 
as in \cite{samadi2018fpca}; 
`default payment' is treated as the binary label. 
We removed individuals with missing values 
and down-sampled the data set from 30,000 
to 20,000. Our preprocessed data sets are published 
at \footnote{https://uwyomachinelearning.github.io/}.

The Communities Crime data set contains 1,993 communities described by 101 informative attributes. 
We treated 
the `fraction of African-American residents' as 
the sensitive feature, and binarized it so that 
a community is 'minority' if the fraction 
is above 0.5 and 'majority' otherwise. Label is the 
`community crime rate', and we binarized it 
into high if the rate is above 0.5 and low otherwise. 

The COMPAS data set contains 18,317 individuals with 40 features (e.g., name, sex, race). We down-sampled the data set to 16,000 instances and 15 numerical features (e.g. name is removed). Similar to  \cite{Chouldechova2017FairPW}, we treated `race' 
as the sensitive feature and `risk of recidivism' 
as the binary label. 

\begin{table*}[t!]
    \def\arraystretch{1.6}
    \setlength{\tabcolsep}{10pt}
    \small   
    \centering
    \noindent\makebox[\textwidth]{
        \begin{tabular}{l||c|c|c|c|c|c}
            \hline
            \multirow{2}{*}{\bf Method} & 
            \multicolumn{2}{c|}{\bf Credit Default} & \multicolumn{2}{c|}{\bf Communities Crime} & \multicolumn{2}{c}{\bf COMPAS} \\
            \cline{2-7}
             & SD & Error & SD & Error & SD & Error \\
            \hline
            FKR-F$^2$E & \textbf{.0021}$\pm$.0017 & 
            .2277$\pm$.0050 & \textbf{.0392}$\pm$.0267 & .1384$\pm$.0125 & \textbf{.0025$\pm$.0018} & .2307$\pm$.0057 \\
            \hline
            FKRR \cite{perez2017fair} & .0079$\pm$.0011 & \textbf{.2001}$\pm$.0054 & .0968$\pm$.0722 & .1208 $\pm$.0054 & .0041$\pm$.0013 & \textbf{.2190}$\pm$.0089 \\
            \hline
            FLR \cite{kamishima2012fair} & .0779$\pm$.0571 & .2412$\pm$.0469 & .0898$\pm$.0971 & .1166$\pm$.0189 & .0408$\pm$.0162 & .2428$\pm$.0917 \\
            \hline
            FRR \cite{McNamara2017ProvablyFR} & .0186$\pm$.0016 & .2914$\pm$.0186 & .3062$\pm$.0452 & \textbf{.1102} $\pm$.0128 & .0182$\pm$.0042 & .2276$\pm$.0040  \\
            \hline
            FPCA \cite{samadi2018fpca} & .1716$\pm$.0149 & .4025$\pm$.0382 & .0859$\pm$.0479 & .1731$\pm$.0089 & .2806$\pm$.0182 & .3204$\pm$.1032 \\
            \hline
        \end{tabular}
    }
    \newline
    \caption{Classification Performance of Different Methods 
    across Different Data Sets. 
    For polynomial kernel, we set degree as 4 and 
    additive coefficient as 0.1. For sigmoid kernel, 
    we used $c = 0.01$ and $\gamma$ as the inverse of 
    feature dimension.
    k is set to $\frac{n}{250}$. 
    }
    \label{tab:classresults}
\end{table*}

\subsection{Experiment Design}

On each data set, we randomly chose 75\% of the 
instances for training and used the rest for testing. 
We evaluated each method over 50 random trials 
and reported its average performance and standard deviation.

We compared the proposed FKR-F$^2$E with the 
existing fair kernel regression \cite{perez2017fair},
and several other non-kernel methods. 
For each compared method, we set its hyper-parameters 
as described in the original paper.

For FKR-F$^2$E, we used polynomial kernel 
on both Credit and Community Crime data sets 
and sigmoid kernel on the COMPAS data set. 
For polynomial kernel, we grid-searched 
its optimal degree in $\{3, 4, 5, 6\}$ 
and optimal additive coefficient in 
$[10^{-3}, 10^{-2}, 10^{-1}, 10^{0}, 10^{1}]$. 
For sigmoid kernel, we grid-searched 
its optimal $c$ among 5 values in the 
logarithmic range of $[10^{-4}, 10^{1}]$,
and we used the default $\gamma$ 
in Scikit-Learn \cite{scikit-learn} 
(i.e., inverse of feature dimension). 
For the ridge regression regularization 
coefficient $\lambda$, we grid-searched 
an optimal value among 6 values in the 
logarithmic range $[10^{-3},10^{2}]$. 

Finally, an important hyper-parameter is 
the number of feature embeddings $k$. 
We experimented with 4 values, namely, 
$\frac{n}{250}$, $\frac{n}{200}$, 
$\frac{n}{150}$ and $\frac{n}{100}$. 
In experiment these values 
yielded good generalization performance on all data sets. 

We evaluated model accuracy using the 
standard classification error (Error), and 
evaluated model fairness using  a 
popular measure called statistical disparity 
(SD) \cite{McNamara2017ProvablyFR}, 
defined as:
\begin{equation}
SD(f,S) 
= |p(f(x)=1 \mid s=1) - p(f(x) = 1 \mid s=0)|.
\end{equation}

Finally, all experiments were run on the Teton Computing Environment at the University of Wyoming's Advanced Research Computing Center (https://doi.org/10.15786/M2FY47), and our FFE implementation is at https://github.com/aokray/FFE.

\subsection{Classification Results and Discussions}

Our classification results are summarized 
in Table \ref{tab:classresults}. 

Our first observation is that FKR-F$^2$E consistently 
achieves lower statistical disparity than the existing 
fair kernel regression method (and other baselines) 
across the three data sets. This implies that  
fair feature embedding is an effective approach 
for learning fair models in kernel space. 

We notice the superior fairness of FKR-F$^2$E 
is not achieved without any cost. In general, it has  
slightly higher prediction error than the existing fair 
kernel regression and other baselines. However, we argue 
the loss of accuracy is small compared with the increase 
of fairness. For example, on the Credit Default data set, 
FKR-F$^2$E lowers prediction disparity by at least 
75\% = (0.0079-0.0021)/0.0079 but only increases 
prediction error by at most 13\% = (0.2277-0.2001)/0.2001.
We thus argue this method has a more efficient 
accuracy-fairness trade-off. 

Finally, we see fair kernel methods generally 
achieve lower statistical disparity than other fair 
learning methods, suggesting their promisingness 
for fair machine learning. 

\subsection{Sensitivity Analysis}

In this section, we examined the performance 
of FKR-F$^2$E on the Communities Crime data set
under different configurations. 

\begin{figure}
    \centering
    \includegraphics[width=0.5\textwidth]{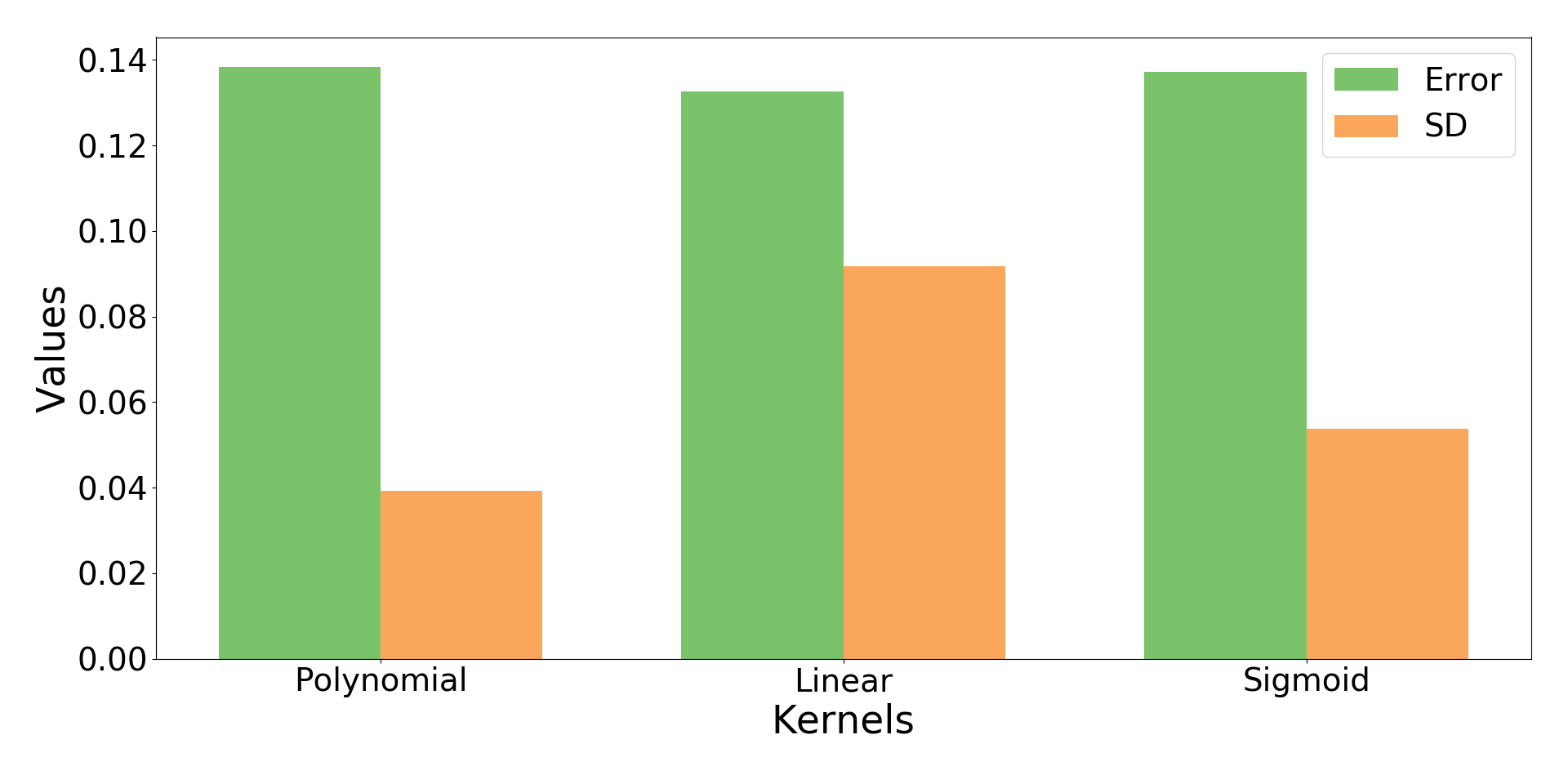}
    \caption{Sensitivity analysis for varying kernels with their approximately optimal number of features selected.}
    \label{fig:kernelsens}
\end{figure}

We first examined its performance with different 
choices of kernel. Results on testing samples 
averaged over 50 random trials are reported 
in Figure \ref{fig:kernelsens}. We see that 
polynomial kernel achieves the highest 
prediction fairness, with slightly higher 
prediction error. Sigmoid kernel is the second 
best, and linear kernel does not give low 
disparity. This supports our hypothesis that 
why fair kernel methods are promising -- 
they construct a complex hypothesis space that 
is more likely to include models with efficient 
fairness-accuracy trade-off. 

\begin{figure}
    \centering
    \includegraphics[width=0.5\textwidth]{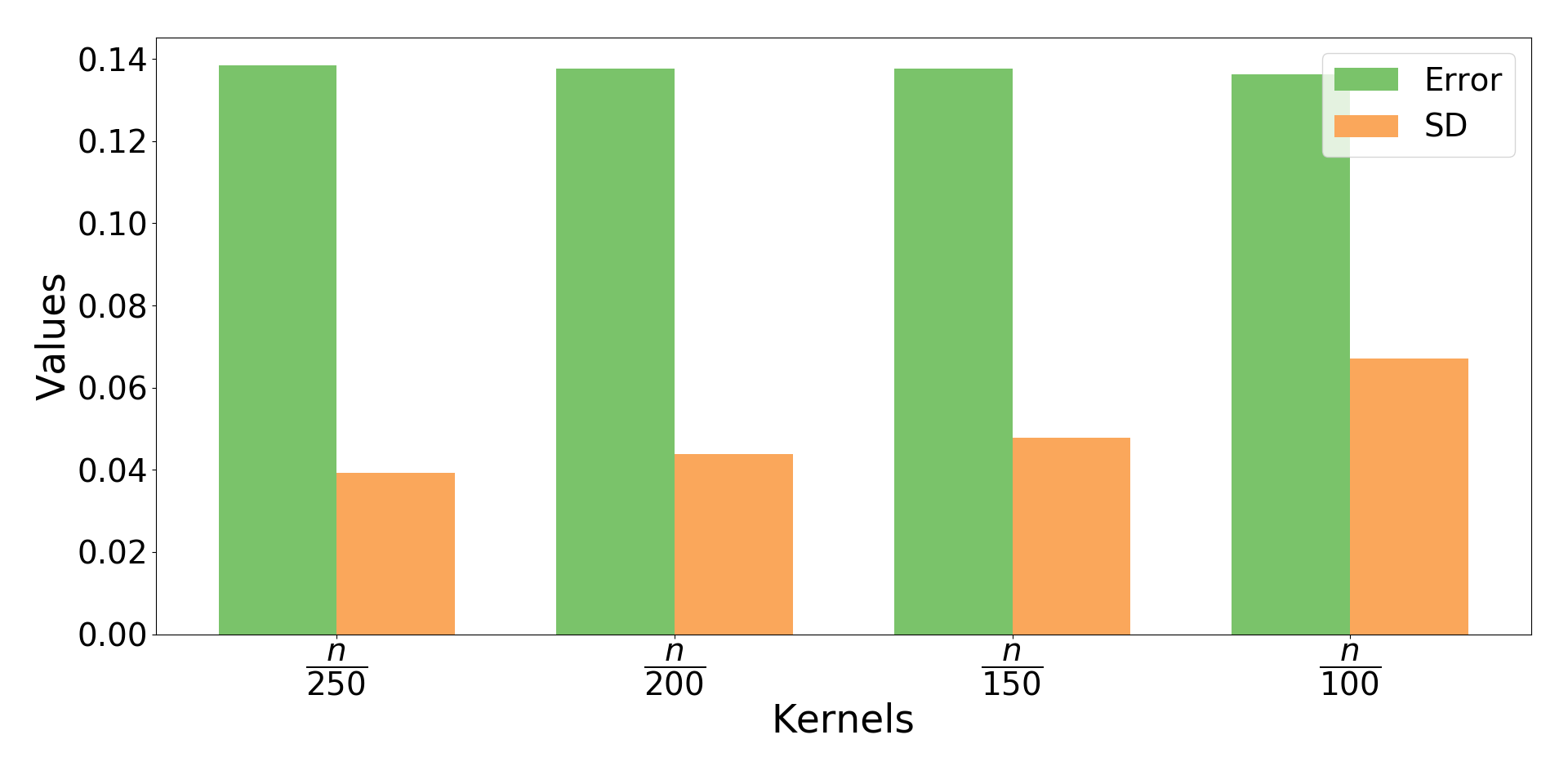}
    \caption{Performance versus $k$.}
    \label{fig:featsens}
\end{figure}

Next, we examined performance with polynomial 
kernel under different $k$ (number of feature
embeddings). Results are shown in Figure
\ref{fig:featsens}. 
We see that smaller $k$ generally leads to 
higher prediction fairness and slightly higher 
prediction error. The former phenomenon implies 
that only the least eigenvectors of problem (\ref{eq:solutioneta}) can effectively minimize 
the mean discrepancy between two groups. 
The latter is easy to understand -- higher 
feature dimension provides more information 
for building an accurate prediction model. 
However, the variation versus $k$ seems 
quite limited, suggesting our method has 
robust classification performance. 

\captionsetup[figure]{skip=0pt}

\section{Conclusion}
In this paper, we propose a novel fair kernel regression 
method FKR-F$^2$E. It first learns a set of fair feature 
embeddings in the kernel space, and then standardly learns 
a prediction model in the embedded space. 
Through experiments across 
three real-world data sets, we show it achieves 
significantly lower bias in prediction 
compared with the state-of-the-art fair kernel 
regression method as well as several non-kernel 
fair learning methods, while sacrificing only 
a small amount of prediction accuracy.

\bibliographystyle{IEEEtran}
\bibliography{reference}

\section{Appendix}

\subsection{Derivation of Eigen-Problem (\ref{eq:solutioneta})}

We will show how to derive (\ref{eq:solutioneta}) 
by solving (\ref{eq:obj}) under constraints 
(\ref{eq:const1}) and (\ref{eq:const2}). 
Recall that $\eta = \sum_{i=1}^{n} 
\alpha_{i} \phi(x_{i})$ where $\alpha_{i}$'s 
are unknown parameters. Rewrite the objective 
in (\ref{eq:obj}) as  
\begin{align}
\begin{split}
    J(\eta) & = \left|\left|\frac{1}{n_u} 
    \sum_{i=1}^{n_u} \langle \phi (x_i), \eta \rangle 
    - \frac{1}{n_p} \sum_{i=n_{u}+1}^{n} 
    \langle \phi(x_i), 
    \eta \rangle\right|\right|^2 \\
     & = \frac{1}{n_u^2} \left( \sum_{i=1}^{n_u}\langle 
     \phi(x_i), \eta \rangle \right)^2 + \frac{1}{n_p^2} 
     \left( \sum_{i=n_{u}+1}^{n} \langle \phi(x_{i}), 
     \eta \rangle\right)^2
       \\ & \quad - \frac{2}{n_u n_p} \left( \sum_{i=1}^{n_u}\langle \phi(x_i), \eta \rangle\right) \left(\sum_{i=n_{u}+1}^{n} \langle \phi(x_{i}, 
       \eta \rangle\right)\\
       & = \frac{\alpha^T K_u^T K_u \alpha}{n_u^2} 
       + \frac{\alpha^T K_p^T K_p \alpha}{n_p^2}
       - \frac{2\alpha^T K_u^T \mathbf{1}_u 
       \mathbf{1}_p^T K_p \alpha}{n_u n_p}\\
       & = \alpha^{T} M \alpha = J(\alpha), 
    \end{split}    
\end{align}
where $M$ is a symmetric matrix defined as 
\begin{equation}
    M = \frac{1}{n_u^2}K_u^T K_u - \frac{2}{n_u n_p}K_u^T \mathbf{1}_u \mathbf{1}_p^T K_p + \frac{1}{n_p^2} K_p^T 
    K_p,  
\end{equation}
matrix $K_u \in \mathbb{R}^{n_u \times n}$ 
has $k(x_{i},x_{j})$ at row $i$ column $j$ 
and matrix $K_p \in \mathbb{R}^{n_p \times n}$  
has $k(x_{n_{u}+i},x_{j})$ at row $i$ column $j$;  
$\mathbbm{1}_u\in \mathbb{R}^{n_u}$ and 
$\mathbbm{1}_p\in \mathbb{R}^{n_p}$ are vectors 
of ones, and $\alpha = [\alpha_{1}, \ldots, \alpha_{n}]$. 

Next, it is easy to verify that constraint 
(\ref{eq:const1}) staisfies 
\begin{equation}
\eta^T\eta = \alpha^{T} K \alpha = 1,  
\end{equation}
where $K \in \mathbb{R}^{n \times n}$ is the 
standard Gram matrix. 

Thus we need to solve 
\begin{equation}
\min_{\alpha} J(\alpha) \quad s.t. \ \alpha^{T} K \alpha = 1.
\end{equation}
The Lagrange function is 
\begin{equation}
\label{eq:lag1}
L(\alpha, \lambda) = J(\alpha) + \lambda (\alpha^{T} K \alpha - 1).
\end{equation}
Setting $\frac{\partial L(\alpha, \lambda)}{\partial \alpha} 
= 0$ and solving for $\alpha$ gives (\ref{eq:solutioneta}). 

\subsection{Derivation of the Solution to (\ref{eq:2ndfunction})}

Here, we show why solution to (\ref{eq:2ndfunction}) 
is also a solution to the eigen-problem (\ref{eq:solutioneta}). 
Let $\alpha$ be the  coefficient vector for the 
first feature embedding $\eta$ (known), and $\alpha'$ be 
the coefficient vector of the second embedding $\eta'$ 
(unknown). 
The new constraint when learning $\eta'$ can be written as 
\begin{equation}
\label{eq:newconstrainta}
    \eta^{T} \eta' = 
    \left( \sum_{i=1}^{n} 
    \alpha_{i} \phi(x_{i}) \right) 
    \left( \sum_{i=1}^{n} 
    \alpha'_{i} \phi(x_{i}) \right) 
    = \alpha^{T} K \alpha' = 0. 
\end{equation}
Thus we need to solve 
\begin{equation}
\min_{\alpha'}\ J(\alpha') \quad s.t.\ 
(\alpha')^{T} K \alpha' = 1, 
\end{equation}
and 
\begin{equation}
(\alpha')^{T} K \alpha = 0.     
\end{equation}
The Lagrange function is 
\begin{equation}
\label{eq:lag2}
L(\alpha', \lambda_{1}, \lambda_{2}) = 
J(\alpha') + \lambda_{1} ((\alpha')^{T} K \alpha' - 1) 
+ \lambda_{2} \alpha^{T} K \alpha'.
\end{equation}
Setting $\frac{\partial L(\alpha', \lambda_{1}, \lambda_{2})}{\partial \alpha'} = 0$ and left-multiplying 
both sides by $\alpha^{T}$, 
\begin{equation}
\alpha^{T} M \alpha' 
- 2\lambda_1 \alpha^{T} K \alpha' - \lambda_2 
    \alpha^{T} K \alpha = 0. 
\end{equation}
Since $\alpha^T K \alpha' = 0$ and 
$\alpha'^T K \alpha' = 1$, we have  
\begin{equation}
\label{eq:Ma}
(\alpha')^{T} M \alpha = \lambda_{2}. 
\end{equation}
Further, from (\ref{eq:solutioneta}) we know $\alpha$ 
is a generalized eigenvector of $M$ satisfying  
$M \alpha = \lambda K \alpha$. Thus (\ref{eq:Ma}) 
becomes 
\begin{equation}
(\alpha')^T M \alpha 
= (\alpha')^T \lambda K \alpha
= \lambda  (\alpha')^T K \alpha 
= 0 = \lambda_{2},  
\end{equation}
where the second equality is due to the new 
constraint (\ref{eq:newconstrainta}). 

Comparing (\ref{eq:lag1}) and (\ref{eq:lag2}), 
with $\lambda_{2}=0$, we see $\alpha$ and 
$\alpha'$ have the same Lagrange function. 
Thus it is easy to show they are solution to 
the same generalized eignproblem (\ref{eq:solutioneta}).  

\end{document}